\documentclass[10pt,twocolumn,letterpaper]{article}

\usepackage[pagenumbers]{iccv} 

\definecolor{iccvblue}{rgb}{0.21,0.49,0.74}
\usepackage[pagebackref,breaklinks,colorlinks,allcolors=iccvblue]{hyperref}

\usepackage{multicol}
\usepackage{hhline}
\usepackage{makecell}
\usepackage{multirow}
\usepackage{booktabs}
\definecolor{lightred}{rgb}{1, 0.2, 0.2}
\definecolor{lightgreen}{rgb}{0.25, 0.75, 0.25}
\usepackage{amssymb}
\usepackage{tabularx}
\usepackage{amsmath}
\usepackage{soul}
\usepackage{colortbl}
\usepackage[accsupp]{axessibility}  

\def\paperID{8573} 
\def\confName{ICCV}
\def\confYear{2025}

\makeatletter
\long\def\@makefntext#1{%
  \parindent 1em%
  \noindent
  \hbox to 0pt{\hss$^{\@thefnmark}$\hspace{0.5em}}#1%
}
\makeatother

\title{UPP: Unified Point-Level Prompting for Robust Point Cloud Analysis}

\author{Zixiang Ai,~~Zhenyu Cui,~~Yuxin Peng,~~Jiahuan Zhou\thanks{Corresponding author: jiahuanzhou@pku.edu.cn}
\\
Wangxuan Institute of Computer Technology,~~Peking University\\
{\tt\small \url{https://azx030512.github.io}, \url{https://zhoujiahuan1991.github.io}}
}

\begin{document}
\maketitle
\begin{abstract}
Pre-trained point cloud analysis models have shown promising advancements in various downstream tasks, yet their effectiveness is typically suffering from low-quality point cloud (i.e., noise and incompleteness), which is a common issue in real scenarios due to casual object occlusions and unsatisfactory data collected by 3D sensors. 
To this end, existing methods focus on enhancing point cloud quality by developing dedicated denoising and completion models. However, due to the isolation between the point cloud enhancement and downstream tasks, these methods fail to work in various real-world domains. In addition, the conflicting objectives between denoising and completing tasks further limit the ensemble paradigm to preserve critical geometric features. To tackle the above challenges, we propose a unified point-level prompting method that reformulates point cloud denoising and completion as a prompting mechanism, enabling robust analysis in a parameter-efficient manner. We start by introducing a Rectification Prompter to adapt to noisy points through the predicted rectification vector prompts, effectively filtering noise while preserving intricate geometric features essential for accurate analysis. Sequentially, we further incorporate a Completion Prompter to generate auxiliary point prompts based on the rectified point clouds, facilitating their robustness and adaptability. Finally, a Shape-Aware Unit module is exploited to efficiently unify and capture the filtered geometric features for the downstream point cloud analysis.
Extensive experiments on four datasets demonstrate the superiority and robustness of our method when handling noisy and incomplete point cloud data against existing state-of-the-art methods. Our code is released at \small{\url{https://github.com/zhoujiahuan1991/ICCV2025-UPP}}.
\end{abstract}
\vspace{-5pt}
\section{Introduction}
\label{sec:introduction}

\begin{figure}[!ht]
  \centering
    \includegraphics[width=\linewidth]{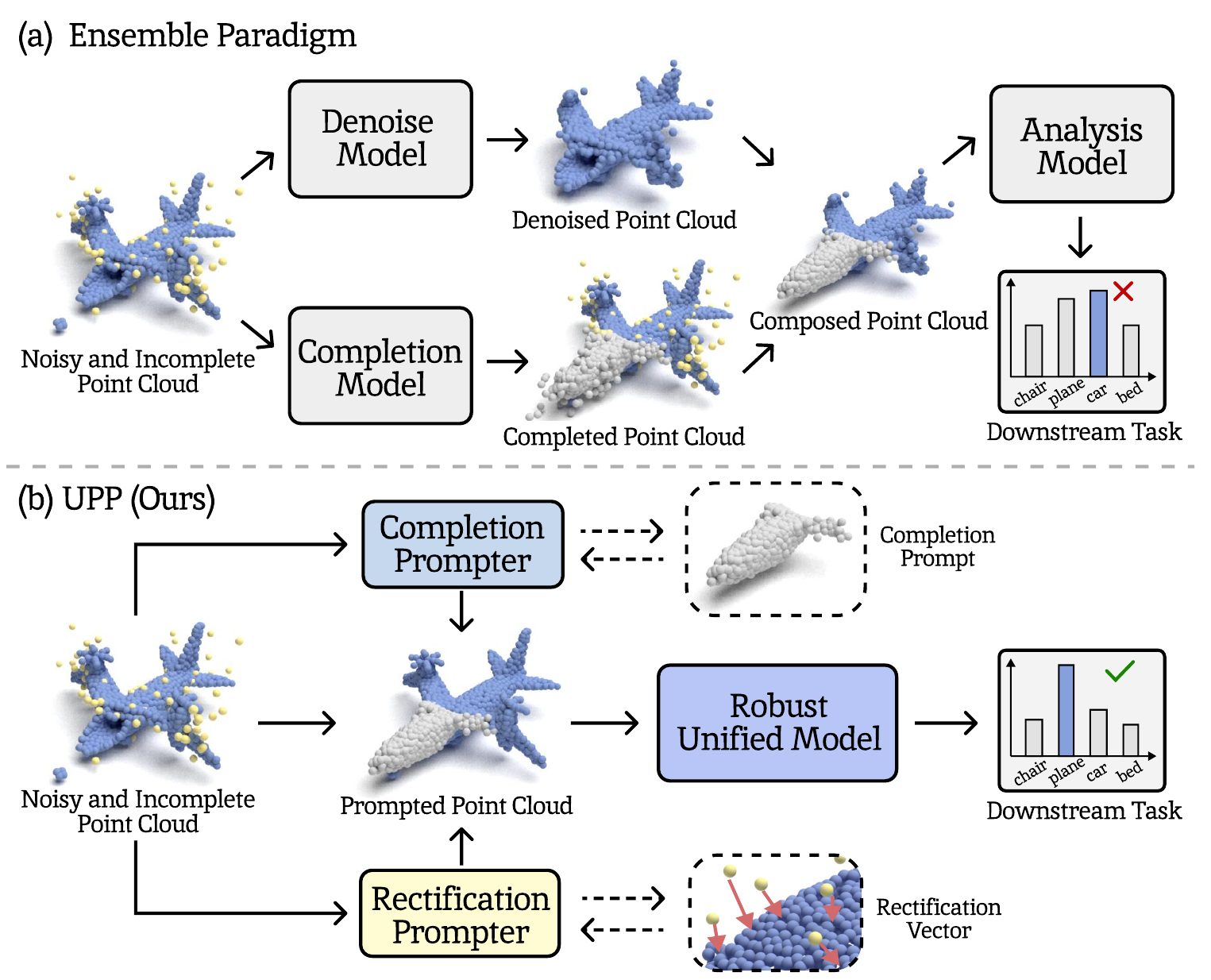}
  \caption{Comparison between (a) the conventional ensemble paradigm utilizing dedicated models and (b) our proposed unified point-level prompting framework. By reformulating denoising and completion tasks as prompting mechanisms tailored for downstream tasks, our approach effectively preserves critical geometric features essential for robust point cloud analysis.}
  \label{fig-short}
  \vspace{-7pt}
\end{figure}

Pre-trained point cloud models have recently achieved significant progress in point cloud analysis, facilitating a wide range of downstream tasks, including 3D object classification~\cite{pang2022PointMAE,zha2023PointFEMAE}, segmentation~\cite{segmentation}, and detection~\cite{detection}. Despite some progress, real-world collected point cloud data typically suffer from substantial noise and incompleteness due to challenges like self-occlusion, reflective surfaces, and the limited sensor resolution~\cite{dmrdenoise,pointr}. These low-quality data critically suppress the performance and reliability of pre-trained models in practical applications, raising an urgent need for effective approaches to ensure real-world scalability and reliability.

To address these challenges, some recent advancements exploited dedicated denoising~\cite{scorebaseddenoise,de2024straightpcf} and completion models~\cite{proxyformer, T-CorresNet2025} and have shown promising results. Specifically, as shown in Figure \ref{fig-short}, denoising models aim to remove redundant point clouds, while completing models focus on adding missing point clouds based on existing point clouds. However, considering the isolation between the point cloud enhancement task and downstream tasks, the performance in downstream tasks typically suffers from the huge gap in task domains. In addition, the simple integration of the above methods fails to handle real-world low-quality point cloud data, aggravating the mutual interference between such two processes, which produces additional missing points during denoising and generates unexpected point clouds in completion due to the domain gap between downstream tasks and pre-training denoising and completion tasks. Consequently, this integration not only diminishes the effectiveness of downstream point cloud analysis but also reduces efficiency due to the complex and cumbersome training pipelines.

To this end, parameter-efficient fine-tuning (PEFT)~\cite{IDPT, tang2024Point-PEFT, ai2025gaprompt} emerges as a promising solution, enabling efficient adaptation of pre-trained point cloud models to various tasks while keeping the backbone parameters frozen.  Unfortunately, most existing PEFT methods~\cite{IDPT, tang2024Point-PEFT, zhou2024DAPT, ai2025gaprompt} ignore the explicit suppression of noise and defects in the input point clouds, resulting in indistinguishable features and suboptimal performance when dealing with low-quality data. As a result, the performance and efficiency of the pre-trained model in downstream tasks are severely degraded.

\begin{figure*}[t]
    \centering
    \includegraphics[width=\linewidth]{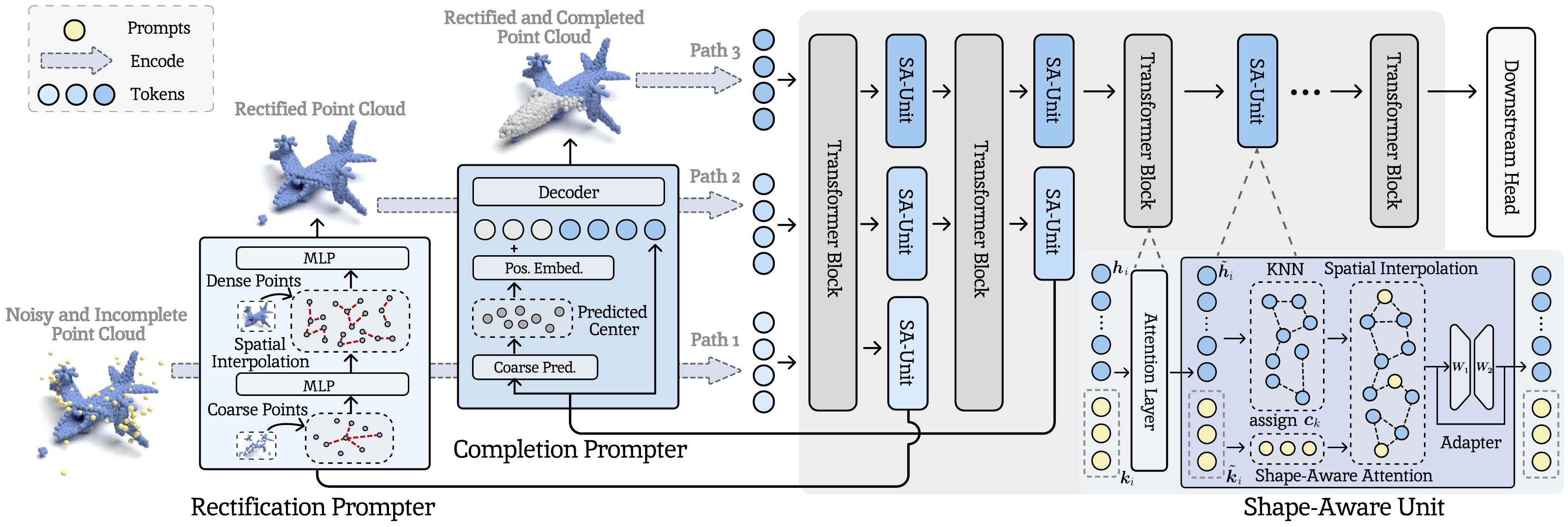}
    \caption{
    Our UPP pipeline processes noisy and incomplete point clouds in a unified paradigm. The input point cloud first passes through shallow blocks to extract features for the Rectification Prompter, adjusting noisy points. Then the rectified point cloud progresses through deeper blocks, where the Completion Prompter predicts missing regions to generate a more complete and representative shape. Finally, features from the enhanced point cloud are aggregated across all blocks to facilitate downstream analysis. Note that we freeze the backbone weights and insert a Shape-Aware Unit (SA-Unit) in each block to efficiently capture essential geometric information, addressing the distinct requirements of both the Rectification and Completion Prompters.
    }
    \label{figure-pipeline}
    \vspace{-9pt}
\end{figure*}

In this paper, we propose Unified Point-level Prompting (UPP), a robust parameter-efficient fine-tuning method that seamlessly unifies downstream point cloud analysis tasks with robust point cloud enhancement, including denoising and completing. To this end, a Rectification Prompter is first proposed to predict and adapt various point cloud noise levels, filtering out noisy points that are irrelevant to downstream tasks, while preserving intricate geometric features crucial for accurate analysis. Besides, a Completion Prompter is further introduced to recover original complete points to recover the destroyed and ignored discriminative information with finer point details. Moreover, to integrate the advantages of the above rectification and completion promoters, a Shape-Aware Unit is further designed to purify the enhanced point cloud structural information in a unified way, strengthening their discriminativeness in downstream tasks with high parameter efficiency. To sum up, our contributions are three-fold:
\begin{itemize}
\item We propose UPP, an end-to-end framework with unified point-level prompts for simultaneous point cloud enhancement and robust analysis, improving model performance on noisy and incomplete data while reducing computational and storage overhead.

\item We introduce three key components, including Rectification Prompter, Completion Prompter, and Shape-Aware Unit, which together enable the model to tackle low-quality point cloud data.

\item Extensive experiments on various benchmarks demonstrate the superior efficiency and effectiveness of UPP, outperforming existing methods in both accuracy and resource utilization.
\end{itemize}
\section{Related Work}
\label{sec:related_work}

\subsection{Point Cloud Pre-training}
Pre-training on 3D datasets has become a prominent research area, particularly with the use of vision transformers~\cite{dosovitskiy2020vit}. Two principal pretext task paradigms have been developed for 3D pre-training: contrastive learning and mask modeling. Methods based on contrastive learning~\cite{ACT_rotate,recon++,pointclipv2} have demonstrated remarkable performance in zero-shot learning, largely due to the inherent power of multi-modality. Mask modeling~\cite{yu2022PointBERT,zha2023PointFEMAE} typically relies on autoencoders to learn the latent features by reconstructing the original input. Credit to the strong characterization capabilities gained from self-supervised learning from large amounts of unlabeled data, the pre-trained model~\cite{pang2022PointMAE,zhang2022PointM2AE} have achieved impressive results across a variety of downstream tasks. However, despite these successes, the effectiveness of pre-trained models is limited when applied to point clouds that are noisy or incomplete, highlighting the need for methods that can enhance robustness in challenging real-world conditions.

\subsection{Point Cloud Enhancement}
Point clouds acquired from scanning devices are often affected by noise and occlusion, compromising downstream tasks such as surface reconstruction and analysis. Enhancing the quality of point clouds, particularly when they are noisy or incomplete, is thus an essential task. Point cloud denoising models have been developed to address this issue and can be categorized into three main types: displacement-based~\cite{pointcleannet}, downsample-upsample~\cite{dmrdenoise}, and score-based methods~\cite{scorebaseddenoise, de2024straightpcf}. Although these methods use different mathematical modeling to estimate noise, they generally consist of a feature extraction module paired with a noise prediction head.
Simultaneously, point cloud completion aims to reconstruct missing regions in partially observed point clouds. PointTr~\cite{pointr} first utilizes transformers to model long-range relationships within the point cloud, enabling accurate completion even in challenging cases with large missing regions. Recent models, such as T-CorresNet~\cite{T-CorresNet2025}, have further improved completion performance by introducing correspondence pooling between query tokens. 

With the rapid development of specialized point cloud denoising and completion models, robust analysis of low-quality point cloud data in downstream tasks has become increasingly feasible. However, this multi-step, ensemble-based paradigm introduces significant computational and storage costs, limiting its practicality for real-time applications. Moreover, the inherent conflict between the objectives of denoising and completion tasks compromises its ability to preserve critical geometric features for real-world analysis tasks. Different from these methods, we reformulate denoising and completion tasks as point-level prompting for downstream tasks, preserving the critical features required for analysis. 

\subsection{Parameter-Efficient Fine-Tuning}
As deep learning technology advances, both the performance and size of models have steadily increased, making full fine-tuning for downstream tasks computationally intensive. To mitigate these challenges, researchers in 2D computer vision have developed various Parameter-Efficient Fine-Tuning (PEFT) methods~\cite{vpt,houlsby2019parameter,jie2023fact,karimi2021compacter,xu2025componential, ai2025vision, liu2025stop, liu2024insvp}. However, due to the inherent sparsity and irregular structure of point clouds, these 2D PEFT methods struggle to generalize effectively to 3D vision tasks. In response, 3D-specific PEFT methods, such as IDPT~\cite{IDPT}, Point-PEFT~\cite{tang2024Point-PEFT}, DAPT~\cite{zhou2024DAPT}, and GAPrompt~\cite{ai2025gaprompt} have been developed to narrow the performance gap with full fine-tuning, achieving efficient adaptation to the unique demands of 3D vision.

However, existing 3D-specific PEFT methods primarily focus on improving representation capacity in the latent feature space with high parameter efficiency.
As a result, the performance of these methods is vulnerable to noisy and incomplete point clouds. This limitation underscores the need for PEFT paradigms that balance between both efficiency and robustness, enabling effective handling of noisy and incomplete point cloud data while remaining representational in downstream analysis.
\section{The Proposed Method}
\label{sec:method}

In this section, we present our Unified Point-Level Prompting (UPP) method for robust point cloud analysis, which consists of the Rectification Prompter, the Completion Prompter, and the Shape-Aware Unit. As shown in Figure \ref{figure-pipeline}, given a pre-trained model's weights, only the inserted modules and the downstream head are trained.

\subsection{Rectification Prompter}
To estimate noise levels per point and enable targeted rectification, we design a Rectification Prompter that effectively filters noise while preserving intricate geometric features essential for analysis. This module is parameter-efficient, utilizing a shared feature extraction backbone with the downstream analysis model, thereby minimizing computational and storage overhead and ensuring seamless integration.

Given a noisy and incomplete point cloud $\boldsymbol{x} \in \mathbb{R}^{S \times 3}$ with $S$ points, we encode it into $L$ tokens $\boldsymbol{h}_0 \in \mathbb{R}^{L \times D}$ along with their positions $\boldsymbol{c} \in \mathbb{R}^{L \times 3}$, where $D$ is the token dimension for the transformer. These tokens are then processed through blocks of the pre-trained model for feature extraction. To satisfy specific feature distribution for noise rectification, we introduce a Shape-Aware Unit following each attention block $\mathcal{H}_i$, tailoring features for the Rectification Prompter as follows:
\begin{equation} 
\boldsymbol{h}_{i+1} = \text{SA-Unit}(\mathcal{H}_i(\boldsymbol{h}_{i}, \boldsymbol{c})), \quad {0 \leq i \leq d_r-1}, 
\end{equation} 
where $d_r$ denotes the number of blocks allocated for the Rectification Prompter.

After obtaining features from $d_r$ blocks, we adopt a coarse-to-fine strategy to propagate features from sparse centers $\boldsymbol{c}$ to dense points $\boldsymbol{x}$. This operation is based on a spatial interpolation denoted as $\mathcal{F}$, described as:
\begin{equation} 
\boldsymbol{f}_{r} = \mathcal{F}(\boldsymbol{h}_{d_r}, \boldsymbol{c}, \boldsymbol{x}) \in \mathbb{R}^{S \times D_r},
\label{interpo1}
\end{equation}
where $\boldsymbol{f}_{r}$ is fine-grained embeddings of each point, with $D_r$ representing the feature dimension and the detail of $\mathcal{F}$ is provided in the appendix. This feature set is then used to estimate noise rectification vector prompts $\boldsymbol{v}_{r} \in \mathbb{R}^{S \times 3} $ through a multi-layer perceptron (MLP), representing both the direction and magnitude of displacement needed for rectification. Points with large $\boldsymbol{v}_{r}$ magnitudes, indicating lower reliability, are masked by leveraging the discrete nature of point clouds. Only points with magnitudes below a threshold $\tau$ are rectified, resulting in a refined point cloud:
\begin{equation} 
    \boldsymbol{x}_{r} = \{\boldsymbol{x} + \boldsymbol{v}_{r}\cdot\alpha \mid \tau > \|\boldsymbol{v}_{r}\|  \} \in \mathbb{R}^{S_r \times 3},
\end{equation}
where $\boldsymbol{x}_{r}$ denotes the rectified points for further processing, $S_r$ is the subset points number and $\alpha$ is a blending factor introduced to prevent over-rectification.

\noindent{\textbf{Objective Function.}} For Rectification Prompter, as shown in Figure \ref{figure-pipeline}, we mix additional noise points $\boldsymbol{n} \in \mathbb{R}^{S_n \times 3}$ into clean points $\boldsymbol{x} \in \mathbb{R}^{S \times 3}$ and predict rectification vectors for each point $i$, denoted as $\boldsymbol{v}_{r}^{i} \in \mathbb{R}^{3}$. The training target for noisy points is the displacement to the clean surface, which can be estimated as the displacement vector to $k$ nearest points in the clean point cloud, denoted as $\boldsymbol{v}_{gt}^{i} \in \mathbb{R}^{3}$ and $k$ is set to 4. For clean points, the target displacement is zero. The loss function is formulated as:
\begin{equation}
\mathcal{L}_{rect} = \frac{1}{S_n}\sum_{i \in \boldsymbol{n}}\| \boldsymbol{v}^i_{r} - \boldsymbol{v}^i_{gt} \|^2 + \frac{1}{S}\sum_{i \in \boldsymbol{x}}\| \boldsymbol{v}^i_{r} \|^2.
\end{equation}

\subsection{Completion Prompter}
The corrected point cloud $\boldsymbol{x}_{r}$ offers enhanced geometric fidelity, enabling the Completion Prompter to accurately infer the overall shape and produce completion point prompts, resulting in a more complete representation. These improvements in point cloud quality empower the analysis model to develop a robust and thorough understanding of the underlying data.

With rectified points $\boldsymbol{x}_{r}$, we resample $L$ local centers $\boldsymbol{c} \in \mathbb{R}^{L \times 3}$ via farthest point sampling, encoding neighboring point patches into tokens $\boldsymbol{h}_0 \in \mathbb{R}^{L \times D}$. These tokens are processed through transformer blocks equipped with Shape-Aware Units tailored for the Completion Prompter.

After processing through $d_c$ blocks, we obtain final tokens $\boldsymbol{h}_{d_c} \in \mathbb{R}^{L \times D}$, which encapsulate rich geometric information about the point cloud instance. Then $\boldsymbol{h}_{d_c}$ is down-projected into concise features and concatenated as a whole feature $\boldsymbol{f}_c$, thereby avoiding information loss typically associated with pooling operations. As shown in Figure \ref{figure-pipeline}, the process is described as follows:
\begin{equation}
\boldsymbol{f}_c = \text{Reshape}(\mathcal{M}(\boldsymbol{h}_{d_c})) \in \mathbb{R}^{D},
\end{equation}
where $\mathcal{M}$ denotes the down-project operation.
Then the $\boldsymbol{f}_c$ is used to predict coarse centers of the missing parts through an MLP head, denoted as $\boldsymbol{c}_m \in \mathbb{R}^{M \times 3}$, where $M$ is the number of predicted coarse points. Notably, MAE-based methods~\cite{pang2022PointMAE, qi2023ReCon, zha2023PointFEMAE} typically use a decoder for point cloud reconstruction, which is often discarded after pre-training. We repurpose its pre-trained weights to reconstruct local patches. This reconstruction process is formalized as follows:
\begin{equation}
\boldsymbol{x}_m = \mathcal{D}([\boldsymbol{h}_{m}+\text{Embed}(\boldsymbol{c}_m), \boldsymbol{h}_{d_c}]),
\end{equation}
where $\boldsymbol{h}_{m} \in \mathbb{R}^{M \times D} $ represents mask tokens, $\boldsymbol{x}_m \in \mathbb{R}^{S_c \times 3} $ are the reconstructed auxiliary point prompts. The $\mathcal{D}$ denotes decoder operation and $[\cdot]$ signifies the concatenation operation. Finally, we combine the rectified partial points with $\boldsymbol{x}_m$ and resample $S$ points using farthest point sampling (FPS) to ensure an even distribution :
\begin{equation}
\boldsymbol{x}_c = \text{FPS}([\boldsymbol{x}_m,\boldsymbol{x}_{r}]) \in \mathbb{R}^{S \times 3} ,
\label{resample}
\end{equation}
where $\boldsymbol{x}_c$ is the final rectified and complete point cloud, rich in representative geometric information.

\noindent{\textbf{Objective Function.}} 
Relying solely on the downstream task loss for supervision often fails to generate meaningful completion prompt points due to insufficient geometric prior knowledge. To address this limitation, we introduce additional supervision for the Completion Prompter by leveraging both the coarse predicted centers and the dense reconstruction. We employ the $\mathcal{L}_1$-norm Chamfer Distance as the metric to evaluate geometric similarity between point clouds. Given two point cloud instances $\mathcal{P}$ and $\mathcal{G}$, the Chamfer Distance function $\mathcal{C}_1(\cdot)$ can be formulated as:
\small
\begin{equation}
\mathcal{C}_1(\mathcal{P}, \mathcal{G}) = \frac{1}{|\mathcal{P}|}\sum_{p \in \mathcal{P}}\min_{g \in \mathcal{G}}\| p - g \| + \frac{1}{|\mathcal{G}|} \sum_{g \in \mathcal{G}} \min_{p \in \mathcal{P}} \| g - p \|,
\end{equation}
\normalsize
where $p$ and $g \in \mathbb{R}^{3} $ represent single point in the instances.

We supervise both the predicted coarse centers $\boldsymbol{c}_m \in \mathbb{R}^{M \times 3}$, the dense completion point prompts $\boldsymbol{x}_m \in \mathbb{R}^{S_c \times 3}$, and the resampled combination $\boldsymbol{x}_c \in \mathbb{R}^{S \times 3}$ of rectified points $\boldsymbol{x}_r$ and $\boldsymbol{x}_m$ from Equation~\ref{resample}. Given ground truth point cloud instance as $\mathcal{P}_{gt}$ and missing point cloud as $\mathcal{P}_m$, the loss for Completion Prompter is formulated as:
\begin{equation}
\mathcal{L}_{comp} = \mathcal{C}_1(\boldsymbol{c}_m, \mathcal{P}_m) + \mathcal{C}_1(\boldsymbol{x}_m, \mathcal{P}_m) + \mathcal{C}_1(\boldsymbol{x}_c, \mathcal{P}_{gt}).
\end{equation}


\subsection{Shape-Aware Unit}
With the enhanced point clouds, the analysis model can effectively capture critical information for downstream tasks. However, directly fine-tuning the pre-trained model to downstream analysis tasks is inefficient in parameters and may lead to catastrophic forgetting of knowledge required by the Rectification Prompter and Completion Prompter. To address this issue, we introduce a Shape-Aware Unit to accommodate the knowledge for point cloud enhancement, thereby capturing intrinsic geometric information essential for downstream tasks. This Shape-Aware Unit is incorporated into each transformer block while keeping the backbone weights frozen, ensuring that only our custom modules are trained during the adaptation process.

For point cloud analysis, given the enhanced points $\boldsymbol{x}_c \in \mathbb{R}^{S \times 3}$, we encode them in into $N$ tokens $\boldsymbol{h}_0 \in \mathbb{R}^{N \times D}$. Then, the feature extraction process is collaboratively performed by the transformer block $\mathcal{H}_i$ and our Shape-Aware Unit. In the $i$-th block, we prepend $K$ prompt tokens $\boldsymbol{k}_i \in \mathbb{R}^{K \times D}$ with the original 3D tokens $\boldsymbol{h}_{i}$. These tokens interact through the self-attention mechanism and are refined by the feed-forward layer:
\begin{equation}
[\tilde{\boldsymbol{k}}_i, \tilde{\boldsymbol{h}}_{i}] = \mathcal{H}_i([\boldsymbol{k}_i,\boldsymbol{h}_{i}])) \in \mathbb{R}^{(K+N) \times D} ,
\end{equation}
where $\tilde{\boldsymbol{k}}_i, \tilde{\boldsymbol{h}}_{i}$ represent the processed prompt and input tokens respectively.

Beyond the feature similarity-based self-attention mechanism, we introduce a Shape-Aware Attention mechanism that builds connections based on spatial distance to enhance robustness. Using the K-nearest neighbor algorithm, we identify the spatial neighboring relationships of 3D tokens based on their positions $\boldsymbol{c}$ and utilize a spatial interpolation function $\mathcal{F}$ to propagate features between local patches.

Furthermore, to incorporate the $K$ prompt tokens into this process, we assign the top $K$ center coordinates of $\boldsymbol{c}$ to $\boldsymbol{k}_i$, denoted as $\boldsymbol{c}_k \in \mathbb{R}^{K \times 3}$. As depicted in Figure~\ref{figure-pipeline}, the procedure can be described as:
\begin{equation}
\hat{\boldsymbol{h}}_{i} = \mathcal{F}([\tilde{\boldsymbol{k}}_i, \tilde{\boldsymbol{h}}_{i}], [\boldsymbol{c}_k, \boldsymbol{c}], \boldsymbol{c}),
\label{interpo2}
\end{equation}
where $\hat{\boldsymbol{h}}_{i}$ is updated 3D tokens and $\mathcal{F}$ is the interpolation function. To prevent feature over-smoothing, a small adapter is employed to adjust the feature distribution:
\begin{equation}
\boldsymbol{h}_{i+1} = W_2\cdot \sigma(W_1(\hat{\boldsymbol{h}}_{i})) + \hat{\boldsymbol{h}}_{i},
\end{equation}
where $W_1 \in \mathbb{R}^{r\times D}$ and $W_2 \in \mathbb{R}^{D\times r}$ respectively denotes the projection weights, $r$ is a hyperparameter controlling the rank, and $\sigma$ is a non-linear activation function. The bias term is omitted for brevity.
After $d$ blocks, we obtain the fully processed tokens $ \boldsymbol{h}_d $ with concentrated geometric information, which are then provided to the downstream task head for further analysis.

\noindent{\textbf{Objective Function.}} 
For point classification tasks with $T$ categories, the task-specific loss is defined as the cross-entropy loss, formulated as:
\begin{equation}
\mathcal{L}_{task}=-\sum_{i=1}^{T}y_i\log(\hat{y}_i),
\end{equation}
where $y_i$ is the ground truth label and $\hat{y}_i$ is the predicted label. 

The overall training loss combines losses for both our point-level promoters and downstream tasks, is formulated as:
\begin{equation}
\mathcal{L}=\mathcal{L}_{rect}+\mathcal{L}_{comp}+\mathcal{L}_{task}.
\end{equation}
This unified loss function ensures that the model simultaneously optimizes point-level prompters and the downstream task, enabling robust and efficient adaptation to real-world scenarios. A staged optimization strategy is employed to further enhance training stability and performance, with detailed implementation provided in the supplementary materials.

\section{Analysis and Discussion}
Given that current pre-trained point cloud models mainly utilize ViT~\cite{dosovitskiy2020vit} architecture, the feature extraction process mainly relies on the self-attention mechanism. 
Given that the raw point clouds are encoded with a lightweight Pointnet~\cite{pointnet}, denoted as $\mathcal{E}$. The encoding process can be formulated as:
\begin{equation}
\boldsymbol{h}_0 = \mathcal{E}(\boldsymbol{x}).
\end{equation}
Then, the attention mechanism with prompt $\boldsymbol{p}_i$ integration can be formally expressed as follows:
\begin{equation}
\mathbf{\hat{o}}_{i} = \text{Attn.}(W_Q\boldsymbol{h}_i, W_K[\boldsymbol{p}_i, \boldsymbol{h}_i], W_V[\boldsymbol{p}_i, \boldsymbol{h}_i]),
\label{attn}
\end{equation}
where $\mathbf{\hat{o}}_{i}$ denote the attention outputs without and with prompt integration. The $W_Q, W_K, W_V$ denote the weights of query, key, and value heads, respectively. 

In our method, the Rectification Prompter and Completion Prompter directly work on $\boldsymbol{x}$ by explicitly moving or appending discrete points, prompting in input data space. As for prompts and adapters introduced in the Shape-Aware Unit, feature distribution can be effectively adjusted within latent token space via the attention mechanism.

Furthermore, given that the self-attention mechanism relies on feature similarity to establish global semantic connections between local patches, this design is susceptible to interference from noisy points. Such interference can cause abrupt changes in local feature similarity, disrupting the model's feature extraction process and ultimately degrading its performance. Therefore, our proposed Shape-Aware Attention mechanism mitigates this issue by constructing attention connections based on spatial distance rather than feature similarity. By leveraging the fact that noisy outlier points are unlikely to alter the spatial neighbouring relationships between local patches, our Shape-Aware Attention enhances the robustness of the original attention mechanism to noise. 

\section{Experiments}
\label{sec:experiments}

\begin{table*}[t]
\renewcommand{\arraystretch}{1.2}
    \centering
    \small
    \begin{tabular}{lccccc}

    \toprule
        \multirow{2}{*}{Method} & \multirow{2}{*}{Reference} & \multirow{2}{*}{Param. (M) $\downarrow$}  & \multirow{2}{*}{FLOPs (G) $\downarrow$}  & \multicolumn{2}{c}{Classification Acc.(\%) $\uparrow$} \\
        \cline{5-6}
         &  &  &  & Noisy ModelNet40 & Noisy ShapeNet55 \\
        \hline
        \multicolumn{6}{c}{\textit{Full Fine-Tuning (FFT)}}  \\
        \hline
        PointNet\textsuperscript{$\dagger$}\cite{pointnet} & CVPR 17   & 3.5 & 0.9 & 74.56 & 71.43  \\ 
        PointASNL\textsuperscript{$\dagger$}\cite{pointasnl} & CVPR 20   & 4.2 & 3.4 & 85.67 & 83.42  \\ 
        PointMLP\textsuperscript{$\dagger$}\cite{pointmlp} & ICLR 22   & 13.2 & 2.0 & 87.88 & 86.72 \\ 
        Point-BERT\textsuperscript{$\dagger$}\cite{yu2022PointBERT} & CVPR 22   & 22.1 & 4.8 & 88.25 & 87.05  \\ 
        Point-MAE\textsuperscript{$\dagger$}\cite{pang2022PointMAE} & ECCV 22   & 22.1 & 4.8 & 89.42 & 88.13  \\ 
        ACT\textsuperscript{$\dagger$}\cite{ACT} & ICLR 23  & 22.1 & 4.8 & 87.24 & 87.39 \\ 
        ReCon\textsuperscript{$\dagger$}\cite{qi2023ReCon} & ICML 23   & 43.6 & 5.3 & 89.67 & 89.01 \\ 
        PointGPT-S\textsuperscript{$\dagger$}\cite{pointgpt} & NeurIPS 23 & 19.5 & 6.1 & 87.48 & 86.35 \\ 
       
        {Point-FEMAE\textsuperscript{$\dagger$}\cite{zha2023PointFEMAE}} & AAAI 24   & 27.4 & 5.0 & 89.59 & 88.63  \\ 
         PCP-MAE\textsuperscript{$\dagger$}\cite{pcpmae} & NeurIPS 24 & 22.1 & 4.8 & 88.21 & 88.24 \\ 
        \hline
        \multicolumn{6}{c}{\textit{Parameter-Efficient Fine-Tuning (PEFT)}}  \\
        \hline
        Point-MAE\textsuperscript{$\dagger$}\cite{pang2022PointMAE} (baseline) & ECCV 22   & 22.1 (100\%) & 4.8 & 89.42 & 88.13  \\ 
        \quad+Point-PEFT\textsuperscript{$\dagger$}\cite{tang2024Point-PEFT} & AAAI 24   & 0.7 (3.2\%) & 7.0  & 87.52\makebox[0pt][c]{\phantom{~~~~~~~~~~~~}\scriptsize{(\color{red} {\textbf{-1.90}}})} & 86.01\makebox[0pt][c]{\phantom{~~~~~~~~~~~~}\scriptsize{(\color{red} {\textbf{-2.12}}})}  \\ 
        \quad+DAPT\textsuperscript{$\dagger$}\cite{zhou2024DAPT} & CVPR 24   & 1.1 (5.0\%) & 5.0  & 86.43\makebox[0pt][c]{\phantom{~~~~~~~~~~~~}\scriptsize{(\color{red} {\textbf{-2.99}}})} & 86.33\makebox[0pt][c]{\phantom{~~~~~~~~~~~~}\scriptsize{(\color{red} {\textbf{-1.80}}})}  \\ 
        \rowcolor{blue!6}\ \ \ \ +UPP (Ours) &  This Paper   & {1.4} ({6.3\%}) & 6.5 & \textbf{92.95}\makebox[0pt][c]{\phantom{~~~~~~~~~~~~}\scriptsize{(\color{blue} {\textbf{+3.53}}})} & \textbf{90.40}\makebox[0pt][c]{\phantom{~~~~~~~~~~~~}\scriptsize{(\color{blue} {\textbf{+2.27}}})}  \\ 
         \hline
         
        {ReCon\textsuperscript{$\dagger$}\cite{qi2023ReCon}} (baseline) & ICML 23 & 43.6 (100\%) & 5.3 & 89.67 & 89.01  \\
        \quad+Point-PEFT\textsuperscript{$\dagger$}\cite{tang2024Point-PEFT} & AAAI 24   & 0.7 (1.6\%) & 7.0  & 88.21\makebox[0pt][c]{\phantom{~~~~~~~~~~~~}\scriptsize{(\color{red} {\textbf{-1.46}}})} & 87.08 \makebox[0pt][c]{\phantom{~~~~~~~~~~~~}\scriptsize{(\color{red} {\textbf{-1.93}}})}  \\ 
        \quad+DAPT\textsuperscript{$\dagger$}\cite{zhou2024DAPT} & CVPR 24   & 1.1 (2.5\%) & 5.0 & 88.41\makebox[0pt][c]{\phantom{~~~~~~~~~~~~}\scriptsize{(\color{red} {\textbf{-1.26}}})} & 86.63\makebox[0pt][c]{\phantom{~~~~~~~~~~~~}\scriptsize{(\color{red} {\textbf{-2.38}}})}  \\
        \rowcolor{blue!6}\quad{+UPP (Ours)} &  This Paper   & {1.4} ({3.2\%}) & 6.5 & \textbf{91.69}\makebox[0pt][c]{\phantom{~~~~~~~~~~~~}\scriptsize{(\color{blue} {\textbf{+2.02}}})} & \textbf{89.68}\makebox[0pt][c]{\phantom{~~~~~~~~~~~~}\scriptsize{(\color{blue} {\textbf{+0.67}}})}  \\ 
        \hline
        
        Point-FEMAE\textsuperscript{$\dagger$}\cite{zha2023PointFEMAE} (baseline) & AAAI 24 & 27.4 (100\%) & 5.0 & 89.59 & 88.63  \\
        \quad+Point-PEFT\textsuperscript{$\dagger$}\cite{tang2024Point-PEFT} & AAAI 24   & 0.7 (2.6\%) & 7.0  & 87.60\makebox[0pt][c]{\phantom{~~~~~~~~~~~~}\scriptsize{(\color{red} {\textbf{-1.99}}})} & 85.16\makebox[0pt][c]{\phantom{~~~~~~~~~~~~}\scriptsize{(\color{red} {\textbf{-3.47}}})}  \\ 
        \quad+DAPT\textsuperscript{$\dagger$}\cite{zhou2024DAPT} & CVPR 24 & 1.1 (4.0\%) & 5.0 & 86.59\makebox[0pt][c]{\phantom{~~~~~~~~~~~~}\scriptsize{(\color{red} {\textbf{-3.00}}})} & 83.45\makebox[0pt][c]{\phantom{~~~~~~~~~~~~}\scriptsize{(\color{red} {\textbf{-5.18}}})}  \\
        
        \rowcolor{blue!6}
        \quad{+UPP (Ours)} & This Paper   & {1.4} ({5.1\%}) & 6.5 & \textbf{91.94}\makebox[0pt][c]{\phantom{~~~~~~~~~~~~}\scriptsize{(\color{blue} {\textbf{+2.35}}})}& \textbf{90.08}\makebox[0pt][c]{\phantom{~~~~~~~~~~~~}\scriptsize{(\color{blue} {\textbf{+1.45}}})}\\
        
        \bottomrule
    \end{tabular}
    \normalsize
    \vspace{-2pt}
    \caption{Classification on Noisy ModelNet40~\cite{shapenet40} and Noisy ShapeNet55~\cite{shapenet55}, including the trainable parameter numbers (Param), computational cost (FLOPs), and overall accuracy. 
    \textsuperscript{$\dagger$} denotes reproduced results using official code. 
    Point cloud classification accuracy without voting is reported.}
    \label{table-modelnet}
    \vspace{-12pt}
\end{table*}

\subsection{Implement Details}
We evaluate the performance of our proposed UPP on the point cloud classification and segmentation tasks. Three representative pre-trained models, Point-MAE~\cite{pang2022PointMAE}, ReCon~\cite{qi2023ReCon}, and Point-FEMAE~\cite{zha2023PointFEMAE} are selected as backbones. 
For benchmarks, we generate noisy and incomplete samples from the synthetic datasets ShapeNet55~\cite{shapenet55} and ModelNet40~\cite{shapenet40} and incomplete samples from the real-world ScanObjectNN~\cite{scanobjectnn} dataset for the inherent noise from sensors.
We train the models on the noisy and incomplete training sets and evaluate them on the standard test dataset.
To ensure a fair comparison, identical data augmentation techniques are applied to each baseline. All experiments are conducted on a single GeForce RTX 4090 GPU. More details on the training and inference processes are available in the supplementary material.

\subsection{Datasets}
\noindent{\textbf{ModelNet40 Dataset.}} ModelNet40~\cite{shapenet40} consists of 12,311 3D CAD models across 40 categories, providing complete, uniform, and noise-free point clouds that serve as ground truth. Following the procedures in PointASNL~\cite{pointasnl} and ScoreDenoise~\cite{scorebaseddenoise}, we add 24 random outlier noise points and 64 surface noise points. Additionally, we use the online cropping method from PoinTr~\cite{pointr} to simulate real-world noise and incompletion scenarios.
To generate the noisy and incomplete point clouds, we first randomly select a viewpoint and remove the 25\% furthest points from that viewpoint. Then, for each instance, we sample 1024 points from the partial ground truth and concatenate the noise points to form the final training point cloud.
Since voting strategy~\cite{voting} is computationally expensive, we focus on reporting the overall accuracy without it.

\noindent{\textbf{ShapeNet55 Dataset.}}
ShapeNet55~\cite{shapenet55} contains about 51,300 unique clean point cloud models across 55 object categories, providing a more challenging classification task due to the complex category distribution. We apply the same noise and incompletion settings as in Noisy ModelNet40, including 25\% missing points, 24 random outlier noise points, and 64 surface noise points.

\noindent{\textbf{ScanObjectNN Dataset.}}
The ScanObjectNN~\cite{scanobjectnn} is a challenging 3D dataset comprising 15K real-world objects across 15 categories. These objects consist of indoor scene data obtained by scanners, containing inherent noise. To avoid the background interference, we select the OBJ\_ONLY split and adopt 25\% incompleteness.


\begin{table}[!ht]
\setlength{\tabcolsep}{4pt} 
\renewcommand{\arraystretch}{1.2}
    \centering
    \small
    \begin{tabular}{l|c|c|c}
    \hline
    Method & Reference& Param.(M)  & Acc.(\%) \\ 
    \hline
    {Point-MAE\textsuperscript{$\dagger$}\cite{pang2022PointMAE}} & ECCV 22 & 22.1 & 88.12    \\ 
    {ReCon\textsuperscript{$\dagger$}\cite{qi2023ReCon}} & ICML 23 & 43.6 & 90.36    \\ 
    {Point-FEMAE\textsuperscript{$\dagger$}\cite{zha2023PointFEMAE}} & AAAI 24 & 27.4 & 90.71    \\ 
    {PCP-MAE\textsuperscript{$\dagger$}\cite{zha2023PointFEMAE}} & NeurIPS 24 & 22.1 & 88.98  \\ 
    \hline
    {Point-FEMAE\textsuperscript{$\dagger$}\cite{zha2023PointFEMAE}} & AAAI 24 & 27.4 & 90.71    \\ 
    {\quad+IDPT\textsuperscript{$\dagger$}\cite{IDPT}} & ICCV 23 & 1.7 & 88.64    \\  
    {\quad+Point-PEFT\textsuperscript{$\dagger$}\cite{tang2024Point-PEFT}} & AAAI 24 & 0.7 & 89.16   \\
    {\quad+DAPT\textsuperscript{$\dagger$}\cite{zhou2024DAPT}} & CVPR 24 & 1.1 & 89.67   \\ 
    \rowcolor{blue!6} {\quad+UPP (Ours)} & This Paper & {1.4} & \textbf{91.39}    \\ 
    \hline
    \end{tabular}
    \normalsize
    \vspace{-2pt}
    \caption{Experiments on real-world dataset ScanObjectNN~\cite{scanobjectnn} with incompleteness and inherent noise.}
    \label{table-scanobject}
\end{table}

\begin{table}[t]
\renewcommand{\arraystretch}{1.1}
\setlength{\tabcolsep}{4pt} 
    \centering
    \small
    \begin{tabular}{c|c|c|c|c}
    \hline
        ~Base~  & Rect. Promp. & Compl. Promp. &  \ \ SA-Unit  \ \ & Acc.(\%) \\
        \hline
        \checkmark&- & - & - & 88.41 \\ 
        \checkmark&\checkmark & - & - & 89.91 \\ 
        \checkmark&\checkmark  & \checkmark & - & 91.41  \\  
        \rowcolor{blue!6} 
        \checkmark& \checkmark & \checkmark & \checkmark & \textbf{92.95}  \\  
    \hline
    \end{tabular}
    
    \normalsize
    \vspace{-2pt}
    \caption{Ablation on effects of each component in our paradigm.}
    \label{table:ablation-on-main-components}
    \vspace{-2pt}
\end{table}

\subsection{Quantitative Analysis}
\noindent{\textbf{Performance on Noisy ModelNet40.}} As shown in Table \ref{table-modelnet}, our method surpasses the state-of-the-art PEFT method DAPT~\cite{zhou2024DAPT} by a large margin due to the robustness to handle low-quality data. Furthermore, our method even surpasses all the fine-tuning of Point-MAE~\cite{pang2022PointMAE}, ReCon\cite{qi2023ReCon}, and Point-FEMAE~\cite{zha2023PointFEMAE} by \textbf{3.53\%}, \textbf{2.02\%}, \textbf{2.35\%} respectively. 
This performance improvement verifies the superiority of reformatting denoising and completion tasks as point-level prompting for robust analysis tasks. 

In terms of efficiency, our approach requires only \textbf{1.4 M} trainable parameters, achieving a reduction of more than \textbf{95\%} in trainable parameters compared to full fine-tuning while introducing little computational cost. This advantage stems from both our Shape-Aware Unit, which effectively captures critical geometric features and unifies diverse enhancement knowledge within a single model, and our point-level prompters for efficiently adapting point clouds within input space.

\noindent{\textbf{Performance on Noisy ShapeNet55.}}
Considering the large data scale and diverse categories of ShapeNet55\cite{shapenet55}, this dataset poses a significant challenge to the representational capabilities of our method. Nevertheless, our approach outperforms the fine-tuning methods, achieving an average improvement of 1.46\% with remarkable parameter efficiency, as shown in Table~\ref{table-modelnet}. This improvement can be attributed to the enhanced representational capability enabled by our PEFT module, the Shape-Aware Unit, and the effectiveness of our point-level prompters in handling low-quality data. Specifically, the Shape-Aware Unit excels at capturing both filtered geometric features and completed detail information, effectively mitigating the interference caused by noise. The above experimental results verify the robustness and efficiency of our method in real-world scenarios with noisy and incomplete point cloud data.

\noindent{\textbf{Performance on Incomplete ScanObjectNN.}}
We further validate the generalizability of our method on real-world scanned object data, as shown in Table~\ref{table-scanobject}. Despite the diverse distribution of noise and fragments in real sensor data compared to simulated noise, our method consistently outperforms other PEFT and fine-tuning approaches. This success is attributed to our unified framework, which integrates point-level prompting with downstream task adaptation, demonstrating strong robustness and adaptability in real-world scenarios. These results highlight the effectiveness of our approach in handling the complexities of real-world point cloud data.

\begin{figure}[t]
    \includegraphics[width=0.95\linewidth]{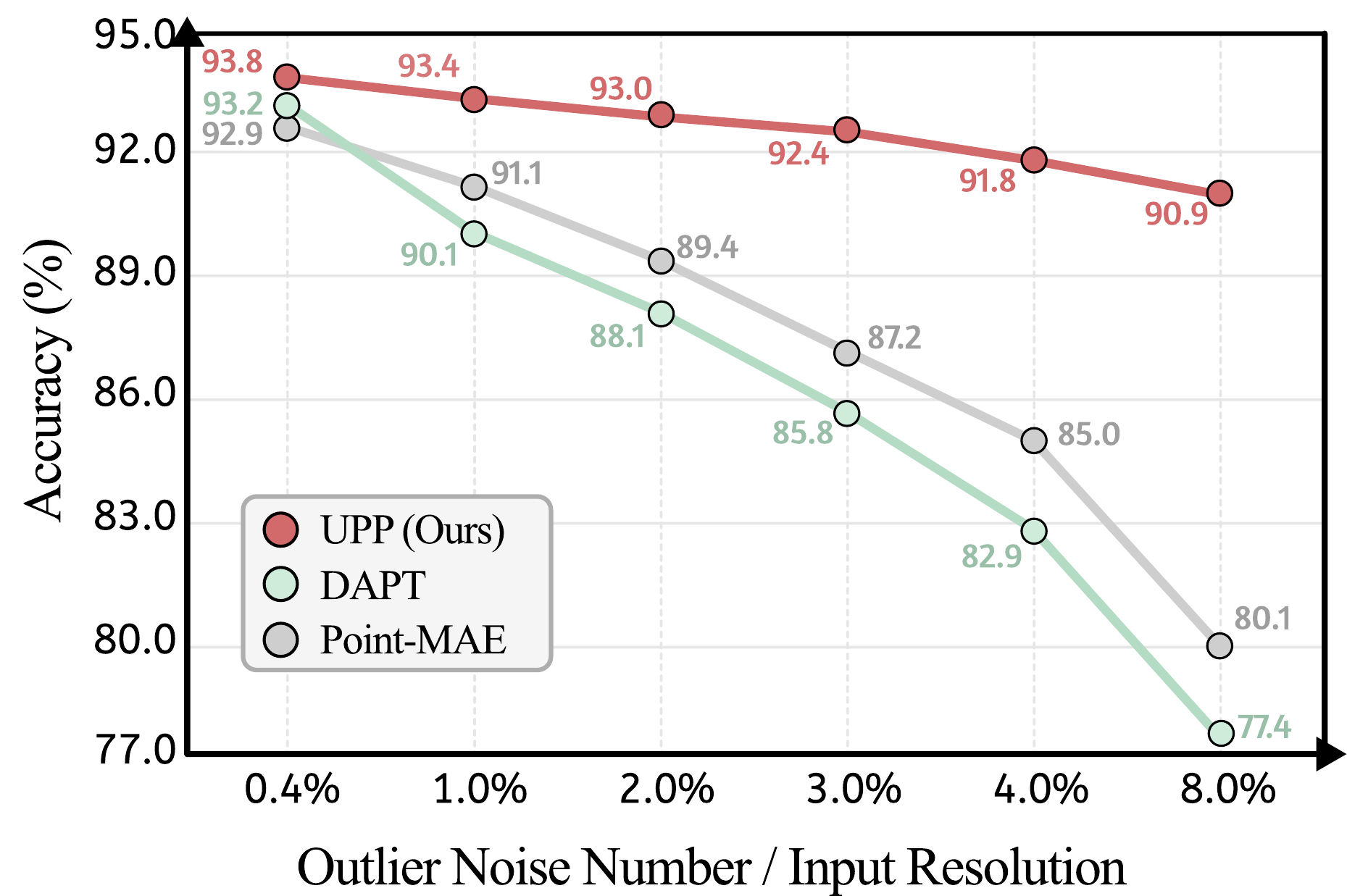}
    \vspace{-2pt}
    \caption{Robustness of our method UPP and other methods~\cite{pang2022PointMAE,zhou2024DAPT} under different outlier noise points number.}
    \label{figure-noise-robust}
    \vspace{-2pt}
\end{figure}

\noindent{\textbf{Robustness to Outlier Noise Levels.}} As shown in Figure ~\ref{figure-noise-robust}, we evaluate the robustness of our method to varying outlier noise levels on ModelNet40, adjusting the number of outliers to simulate different noise intensities. The curves indicate that as the outliers number increases, both the baseline model Point-MAE~\cite{pang2022PointMAE} and PEFT method DAPT~\cite{zhou2024DAPT} struggle to capture essential geometric information, resulting in rapid performance degradation. Notably, our classification accuracy remains competitive with an outlier ratio under 2\%, demonstrating the efficacy of our point-level prompters in enhancing point clouds, providing filtered geometric features and comprehensive shape information for downstream analysis.

    

\subsection{Ablation Studies}

In this section, we conduct extensive experiments on Noisy ModelNet40 to evaluate the impact of each component on our method. We adopt pre-trained Point-MAE~\cite{pang2022PointMAE} as the backbone for ablation. More ablation experiments can be found in the supplementary material.

\noindent{\textbf{Effectiveness of Each Component.}} As shown in Table \ref{table:ablation-on-main-components}, we sequentially add our Rectification Prompter, Completion Prompter, and Shape-Aware Unit (SA-Unit) to the base linear probing of pre-trained backbone method to evaluate their contributions. When the rectification prompter is employed, the accuracy of our method is improved by 1.50\%, demonstrating its ability to effectively filter noise while preserving complex geometric features essential for analysis. The combination of both the Rectification Prompter and the Completion Prompter further boosts performance, achieving an additional improvement of 1.50\%. This validates that our point-level prompting mechanism enriches the discriminative geometric features, providing a more comprehensive representation of the point cloud. Finally, the introduction of the SA-Unit improves the performance by 1.54\%, which can be attributed to its Shape-Aware Attention design. This mechanism facilitates interaction between prompt and input tokens through both self-attention and spatial distance-based attention, effectively capturing critical shape information and further enhancing the model's robustness and accuracy.

\begin{figure}[t]
    \centering
    \includegraphics[width=0.95\linewidth]{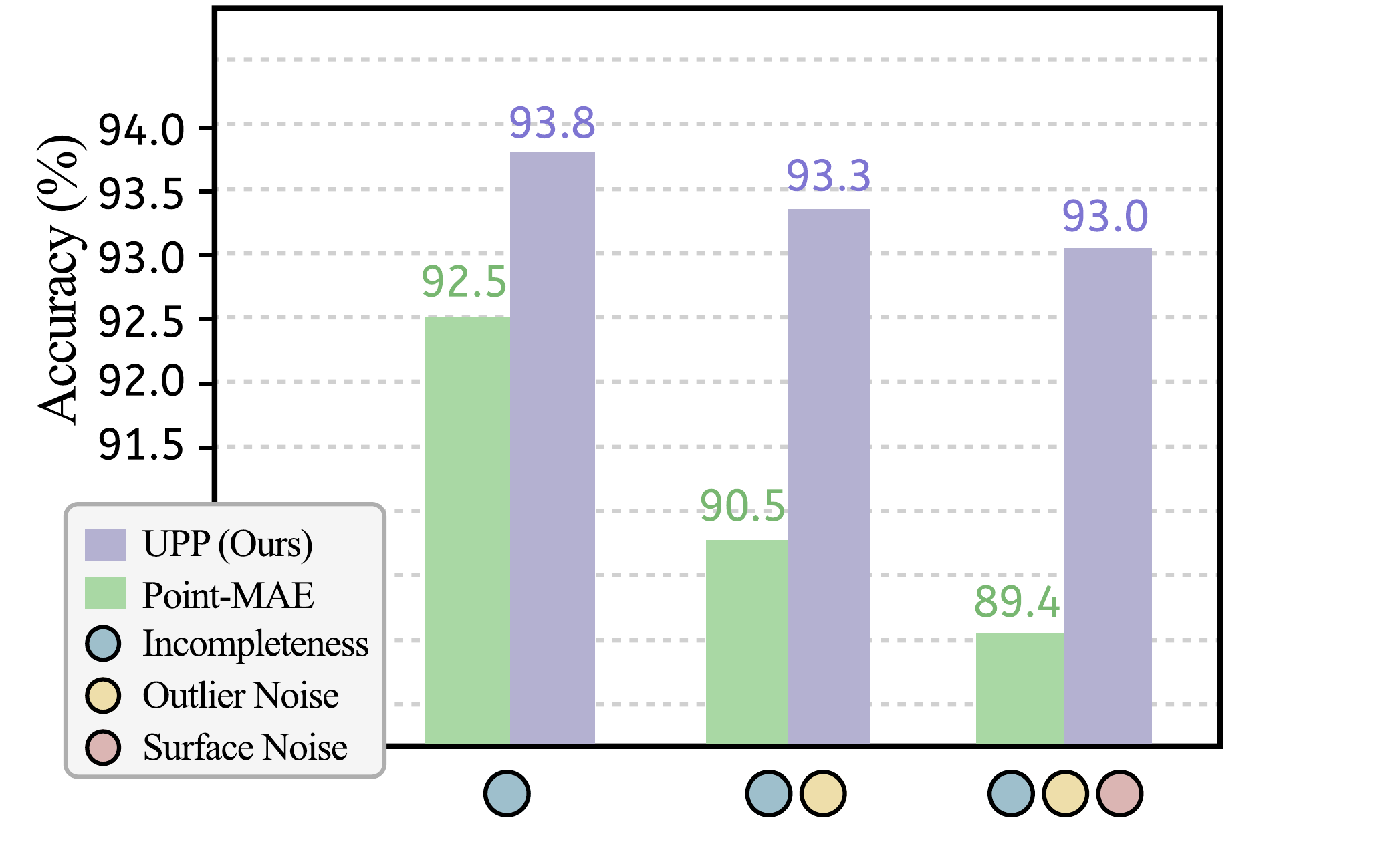}
    \caption{Impairment of model performance by different forms of point cloud noise or incompleteness.}
    \label{figure:ablation-on-different-noise}
    \vspace{-12pt}
\end{figure}

\noindent{\textbf{Impairment of Different Noise Types.}} As demonstrated in Figure~\ref{figure:ablation-on-different-noise}, we conduct experiments on the impairment of different kinds of noise to analysis tasks. For fine-tuning of Point-MAE\cite{pang2022PointMAE}, its performance steadily declines with a combination of 25\% incompleteness, 24 outlier noise points, and 64 surface noise points. 
In contrast, our method maintains downstream classification accuracy with minimal degradation, exhibiting robustness to different kinds of noise. It is attributed to our point-level prompts effectively filtering noise while preserving complex geometric features and generating more complete representations, thus benefiting the downstream analysis.

\begin{figure}[t]
  \centering
    \includegraphics[width=\linewidth]{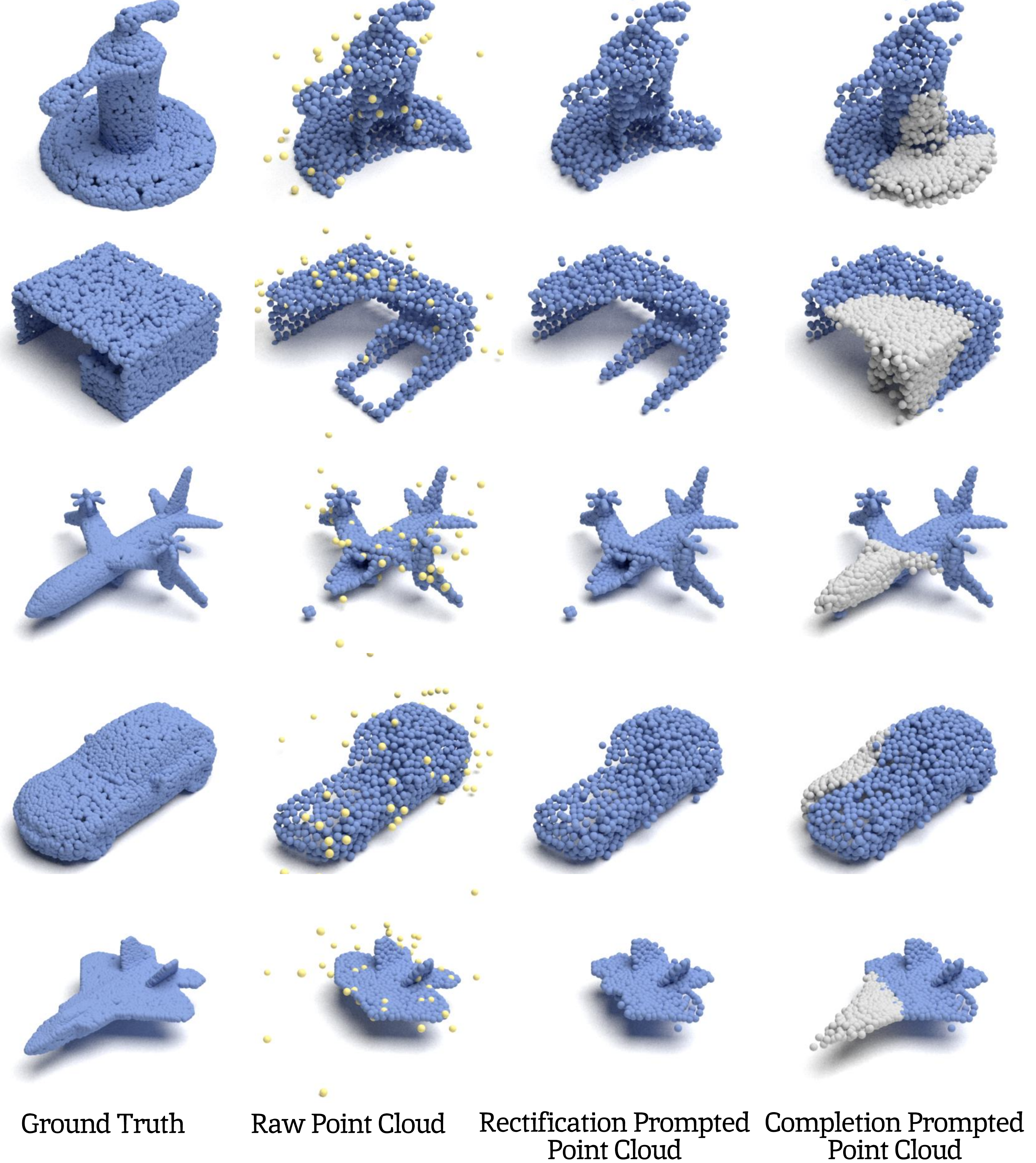}
  \caption{Visualization of Noisy ModelNet40 dataset. Our Rectification Prompter and Completion Prompter explicitly prompt the analysis at point levels.}
  \label{fig:visualization}
  \vspace{-10pt}
\end{figure}

\subsection{Visualization} 
Figure \ref{fig:visualization} depicts the visualization of the noisy and incomplete point cloud and corresponding rectified and completed point clouds. It can be seen that the Rectification Prompter exactly corrects most noisy points without hurting the intricate geometry structures, attributed to accurate point cloud feature extraction provided by the Shape-Aware Unit. And Completion Prompter effectively predicts the missing parts, providing more complete shapes for feature extraction for downstream tasks.

\section{Conclusion}
\label{sec:conclusion}

In this paper, we introduce Unified Point-level Prompting (UPP), an end-to-end framework that reformulates point cloud denoising and completion as a prompting mechanism, enabling robust analysis in a parameter-efficient manner. We demonstrate that unifying point-level enhancement with the analysis model significantly improves downstream task performance while introducing minimal computational overhead. To achieve this, we design a Rectification Prompter and a Completion Prompter to provide point-level prompts, alongside a Shape-Aware Unit that integrates diverse enhancement knowledge requirements with parameter-efficient representational capabilities. 
Our proposed UPP approach is empirically validated to be robust under various low-quality point cloud conditions while maintaining high parameter efficiency. This framework not only enhances the geometric fidelity of point clouds but also ensures seamless adaptation to downstream tasks, making it a practical solution for real-world applications.

\section*{Acknowledgments}
This work was supported by the National Natural Science Foundation of China (62376011) and the National Key R\&D Program of China (2024YFA1410000).

{
    \small
    \bibliographystyle{ieeenat_fullname}
    \bibliography{UPP}
}
\clearpage
\setcounter{page}{1}
\maketitlesupplementary

\section*{Training Detail}
\noindent\textbf{Downstream tasks in noisy and incomplete condition.} 
In our experiments, we train the downstream classifiers under noisy conditions. For fair comparisons, identical hyper-parameters and training strategies are applied across fine-tuning and proposed methods, following the pioneering work Point-MAE~\cite{pang2022PointMAE}, as shown in Table \ref{train-details}. For example, when fine-tuning on Noisy ModelNet40~\cite{shapenet40}, the training process spans 300 epochs, using a cosine learning rate scheduler~\cite{loshchilov2022sgdr} that starts at 0.0005, with a 10-epoch warm-up period. The AdamW optimizer~\cite{loshchilovdecoupled} is employed to facilitate optimization. To evaluate performance, we utilize the overall accuracy metric, comparing the model's predictions on the clean test set. 

All of our experiments across the four datasets adhere to the settings outlined in Table~\ref{train-details}, with the exception of the ScanObjectNN dataset. For ScanObjectNN, we set the point number to 2048 and adopt 128 patches to better accommodate the characteristics of real-world scanned data, following previous works~\cite{pang2022PointMAE,zha2023PointFEMAE}.

\noindent\textbf{Parameter-Efficient Fine-tuning Settings.} 
As a parameter-efficient fine-tuning method, we merely train our inserted modules with pre-trained backbone weights frozen. Following the approach of DAPT~\cite{zhou2024DAPT}, we load pre-trained weights into a Point-MAE~\cite{pang2022PointMAE} model for efficient fine-tuning, excluding residual components for consistency. 
Notably, ReCon~\cite{qi2023ReCon} and Point-FEMAE~\cite{zha2023PointFEMAE} extend Point-MAE~\cite{pang2022PointMAE} with additional modules. We drop these parameters, thus leading to a slight saving of FLOPs. All experiments are implemented using PyTorch version 1.13.1 and conducted on a single GeForce RTX 4090 GPU.

\begin{table}[!ht]
    \centering
    \small
    \setlength{\extrarowheight}{2pt} 
    \begin{tabular}{lcc}
     \toprule
        Task & Classification & Segmentation \\   
        \midrule
        Optimizer & AdamW & AdamW \\ 
        Learning rate & 0.0005 & 0.0002 \\ 
        Weight decay & 0.05 & 0.05 \\ 
        Scheduler  & cosine & cosine\\ 
        Training epochs & 300 & 300 \\ 
        Warmup epochs & 10 & 10 \\ 
        Batch size & 32 & 32 \\ 
        Outliers number & 24 & 24 \\ 
        Surface noise number & 64 & 64 \\ 
        Shape missing rate & 25\% & 25\% \\
        Points number & 1024 & 2048 \\  
        Patches number & 64 & 128 \\ 
        Patch size & 32 & 32 \\ 
        \bottomrule
    \end{tabular}
    \caption{Training details for downstream classification and segmentation tasks in noise condition.}
    \label{train-details}
\end{table}

\begin{table}[!ht]
\setlength{\tabcolsep}{2pt} 
\renewcommand{\arraystretch}{1.3}
    \centering
    \small
    \begin{tabular}{l|c|c|c}
    \hline
        Method & Param.(M)  & Cls. mIoU(\%) & Inst. mIoU(\%) \\ 
        \hline
        
        {\textcolor{gray}{Point-MAE}\cite{pang2022PointMAE}}  & \textcolor{gray}{27.06} & \textcolor{gray}{83.3} & \textcolor{gray}{85.6}    \\ 
        {\quad+Point-PEFT\cite{IDPT}} & 5.62 & 80.5 & 83.1  \\  
        {\quad+DAPT\cite{zhou2024DAPT}} & 5.65 & 80.9 & 83.7    \\ 
        \rowcolor{blue!6} 
        {\quad+UPP (Ours)} & 6.43 & \textbf{82.2} & \textbf{84.4}    \\
        \hline
        {\textcolor{gray}{Point-FEMAE}\cite{zha2023PointFEMAE}} & \textcolor{gray}{27.06} & \textcolor{gray}{83.5} & \textcolor{gray}{85.9}  \\ 
        {\quad+Point-PEFT\cite{IDPT}} & 5.62 & 80.7 & 83.9  \\  
        {\quad+DAPT\cite{zhou2024DAPT}} & 5.65 & 81.3 &  84.1  \\ 
        
        \rowcolor{blue!6} 
        {\quad+UPP (Ours)} & 6.43 & \textbf{82.5} & \textbf{84.8}    \\ 
    \hline
    \end{tabular}
    \normalsize
    \vspace{-2pt}
    \caption{Point cloud part segmentation experiment results on ShapeNetPart~\cite{shapenetpart} dataset under noisy and incomplete setting.}
    \label{table-segmentation}
    \vspace{-2pt}
\end{table}

\noindent\textbf{Staged Optimization Strategy.} 
While adapting to downstream tasks, we impose additional objective loss functions to regularize our point-level promoters, the Rectification Prompter and Completion Prompter. During training, we adopt a staged optimization strategy to avoid randomly initialized prompt points disrupting the training of downstream tasks.
We add 50 epochs to optimize the point-level promoters, in which the former 20 epochs optimize both the Rectification Prompter and Completion Prompter, and the later 30 epochs optimize only the Completion Prompter. During the downstream adaptation process, we optionally enable the training of the two point-level promoters with the Shape-Aware Unit when the learning rate narrows to 0.0001.
To simulate real-world noise and incompletion, we introduce additional outlier points and apply random cropping, ranging from 25\% to 50\%, to create labeled data pairs for supervision. During training, the backbone weights are frozen, and only the Rectification Prompter, Completion Prompter, and their associated Shape-Aware Unit modules are optimized.

\begin{table}[!ht]
\setlength{\tabcolsep}{4pt} 
\renewcommand{\arraystretch}{1.1}
    \centering
    \small
    \begin{tabular}{l|c|c|c}
    \hline
        Method & Reference& Param.(M)  & Acc.(\%) \\ 
        \hline
        {Point-FEMAE\cite{pang2022PointMAE}} & AAAI 24 & 27.4 & 94.0    \\ 
        {Linear Probing} & - & 0.3 & 91.9    \\ 
        \hline
        {VPT\cite{vpt}} & ECCV 22 & 0.4 & 92.6    \\ 
        {Adapter\cite{adaptformer}} & NeurIPS 22 & 0.9 & 92.4    \\ 
        {LoRA\cite{hu2021lora}} & ICLR 22 & 0.9 & 92.3    \\ 
        {IDPT\cite{IDPT}} & ICCV 23 & 1.7 & 93.4    \\  
        {Point-PEFT\cite{tang2024Point-PEFT}} & AAAI 24 & 0.7 & 94.0 \\ 
        {DAPT\cite{zhou2024DAPT}} & CVPR 24 & 1.1 & 93.2    \\ 
        \hline
        \rowcolor{blue!6} {SA-Unit (Ours)} & This Paper & \textbf{0.6} & \textbf{94.2}    \\ 
    \hline
    \end{tabular}
    \normalsize
    \caption{Comparison with other PEFT methods on clean ModelNet40~\cite{shapenet40} dataset. Our method only utilizes the PEFT module, Shape-Aware Unit (SA-Unit). Classification accuracy without voting is reported. All methods adopt Point-FEMAE as the backbone.}
    \label{table-ablation-saunit}
\end{table}

\section*{Additional Experiments}
\noindent{\textbf{Segmentation Experiments on Noisy ShapeNetPart.}}
ShapeNetPart~\cite{shapenetpart} includes 16,881 samples across 16 categories for the object-level part segmentation task. It is challenging to accurately recognize class labels for each point within point cloud instances. Furthermore, we add additional simulated noise points and incompleteness, which are detailed in Table~\ref{train-details}. 

As shown in Table~\ref{table-segmentation}, our method outperforms other state-of-the-art PEFT approaches, such as Point-PEFT~\cite{tang2024Point-PEFT} and DAPT~\cite{zhou2024DAPT}, on both pre-trained Point-MAE and Point-FEMAE backbones. This success verifies our method's superior robustness to low-quality data and validates its generalizability across diverse downstream tasks. However, we observe that it remains challenging to surpass the performance of full fine-tuning methods in fine-grained analysis tasks like part segmentation, which require substantial model capacity to memorize the training data distribution. Additionally, current PEFT methods, including ours, exhibit greater susceptibility to noise and incompleteness compared to full fine-tuning.

It is worth noting that the majority of trainable parameters in our framework originate from the large downstream task head, highlighting the efficiency of our approach in minimizing additional parameter overhead while maintaining competitive performance. 

\noindent{\textbf{Comparison with Other PEFT Methods.}} As shown in Table \ref{table-ablation-saunit}, we present classification results on the clean ModelNet40 dataset and compare our Shape-Aware Unit with other PEFT approaches~\cite{vpt, adaptformer, hu2021lora,IDPT,tang2024Point-PEFT,zhou2024DAPT}. Since other methods struggle to tackle low-quality point clouds, we ensure a fair comparison by applying no noise or incompletion settings. Despite these adjustments, our approach achieves the highest accuracy of \textbf{94.2\%}, outperforming both the state-of-the-art PEFT method DAPT~\cite{zhou2024DAPT} and the full fine-tuning. This success is attributed to the effective interaction of the feature similarity-based self-attention mechanism and spatial distance-based Shape-Aware Attention, capturing critical shape information. These results highlight the adaptability and potential to serve as a general 3D PEFT method.

\noindent{\textbf{Impact of Different Prompting Order.}} The order of point-level prompting is a critical factor influencing the performance of our method. As shown in Table~\ref{table:ablation-on-orders}, we compare the impact of different prompting orders. Our results indicate that UPP achieves the highest performance of 92.95\% when the Rectification Prompter is applied first. This suggests that reducing noise levels forms a solid foundation for accurate point cloud understanding, which is essential for both completion prompting and analysis. Intuitively, performing both completion and rectification concurrently could offer better computational parallelism. However, this approach yields only marginal performance improvements. This is because the Completion Prompter relies on the Rectification Prompter to rectify noisy points, enabling filtered features of the point cloud. Interestingly, performing completion before rectification results in improved performance than concurrent, as the Rectification Prompter helps to correct artifacts introduced by low-quality completion. Based on these empirical findings, we adopt the rectification first strategy in our method.

\begin{table}[t] 
\setlength{\tabcolsep}{6pt} 
\renewcommand{\arraystretch}{1.2}
    \centering
    \small
    \begin{tabular}{c|c|c|c}
    \hline
         Concurrently   & Complete First &   Rectify First  & Acc.(\%) \\
        \hline
        \checkmark & - & - & 90.76 \\ 
        -  & \checkmark & - & 91.18  \\  
        \rowcolor{blue!6} -  & - & \checkmark & \textbf{92.95}  \\  
    \hline
    \end{tabular}
    \normalsize
    \caption{Abaltion on point-level prompting order.}
    \label{table:ablation-on-orders}
\end{table}

\begin{table}[t]
    \centering
    \small
    \setlength{\extrarowheight}{2pt} 
    \begin{tabular}{c|c|c}
     \toprule
        Rect. Prompter & Compl. Prompter & Shape-Aware Unit \\   
        \midrule
        0.148M & 0.370M & 0.028M \\ 
        \bottomrule
    \end{tabular}
    \caption{Parameters of each component in our UPP.}
    \label{component-parameters}
    \vspace{-10pt}
\end{table}

\subsection*{Parameters Efficiency}
Our UPP paradigm employs only 1.4 M trainable parameters and requires 6.1 G FLOPs, significantly reducing computational costs compared to the ensemble paradigm while achieving superior performance. This efficiency is attributed to our compact module design and the progressive extraction of point cloud features from shallow to deep layers.

The enhancement in parameter efficiency arises from the insight that, in the ensemble paradigm, the denoising, completion, and analysis models each include dedicated feature extraction modules designed for task-specific knowledge. By contrast, our approach leverages a unified pre-trained backbone for robust feature extraction. Lightweight Shape-Aware Unit modules are then employed to adaptively adjust feature distributions for specific tasks. This unified design substantially improves the efficiency of both the total parameter count and the trainable parameters, achieving a balance between performance and resource utilization, as detailed in Table~\ref{component-parameters}.

\section*{Implementation Detail}
\subsection*{Spatial Interpolation}
We provide detailed formulations for the spatial interpolation operation $\mathcal{F}(\cdot)$ utilized in Equation~\ref{interpo1} and Equation~\ref{interpo2}.

Given a set of center points with coordinates $\{c_i\} \in \mathbb{R}^{C \times 3}$, where $i = 1, \dots, C$, and the corresponding features $\{f(c_i)\}$, the objective of the Propagation operation is to compute the features of a neighboring point $x \in \mathbb{R}^3$ using spatial interpolation. The resulting feature $f(x)$ is derived from $x$, the coordinates $\{c_i\}$, and the features $\{f(c_i)\}$, demonstrated as:
\begin{equation}
    f(x) = \mathcal{F}(\{f(c_i)\}, \{c_i\}, x)
\end{equation}
First, we compute the Euclidean distance from $x$ to each center point $c_i$:
\begin{equation}
    d(x,c_i) = \|x-c_i\| .
\end{equation}
Next, we calculate the weight by taking the inverse of the spatial distance:
\begin{equation}
    w(x,c_i)=\frac{1}{d(x,c_i)^p},
\end{equation}
where $p$ is typically set to 2.

This results in a set of weights ${w(x,c_i)}$ for $i = 1, \dots, C$. We then select only the top-$K$ weights for interpolation:
\begin{equation}
    \{w(x,c_j)\} = \text{Top-}K(\{w(x,c_i)\}),
\end{equation}
where $j = 1, \dots, K$, and $K$ is typically set to 6.
Subsequently, the interpolation of features is based on the weights, formulated as:
\begin{equation}
    f(x) = \frac{\sum_{j=1}^{K}w(x,c_j)f(c_j)}{\sum_{j=1}^{K}w(x,c_j)}.
\end{equation}
Finally, this procedure is repeated for each neighboring point to obtain their features for further utilization.
The Propagation operation effectively transfers and aggregates features by leveraging spatial relationships, enabling robust and efficient feature refinement. This mechanism is particularly suited for point clouds, where irregular data distribution necessitates dynamic interpolation based on spatial distances.

\end{document}